\documentclass[runningheads]{llncs}
\usepackage[T1]{fontenc}
\usepackage{graphicx}
\usepackage{color}
\usepackage{amsmath}
\usepackage{subcaption}
\usepackage{amssymb}
\usepackage{algorithm}
\usepackage{algpseudocode}
\usepackage{tabularx}
\usepackage{float} 
\usepackage{hyperref}

\begin{document}
\title{Lightweight Defense Against Adversarial Attacks in Time Series Classification}
\titlerunning{Lightweight Defense in TSC}
%
\author{Yi Han\inst{1}\orcidID{0009-0001-7296-2562}}
\authorrunning{Y. Han}
%
\institute{Independent Researcher, Australia \\
\email{yi.han@adelaide.edu.au}}
\maketitle              
\begin{abstract}
As time series classification (TSC) gains prominence, ensuring robust TSC models against adversarial attacks is crucial. While adversarial defense is well-studied in Computer Vision (CV), the TSC field has primarily relied on adversarial training (AT), which is computationally expensive. In this paper, five data augmentation-based defense methods tailored for time series are developed, with the most computationally intensive method among them increasing the computational resources by only 14.07\% compared to the original TSC model. Moreover, the deployment process for these methods is straightforward. By leveraging these advantages of our methods, we create two combined methods. One of these methods is an ensemble of all the proposed techniques, which not only provides better defense performance than PGD-based AT but also enhances the generalization ability of TSC models. Moreover, the computational resources required for our ensemble are less than one-third of those required for PGD-based AT. These methods advance robust TSC in data mining. Furthermore, as foundation models are increasingly explored for time series feature learning, our work provides insights into integrating data augmentation-based adversarial defense with large-scale pre-trained models in future research. This version of the contribution has been accepted for publication, after peer review (when applicable), but is not the Version of Record and does not reflect post-acceptance improvements or any corrections. The Version of Record is available online at: http://dx.doi.org/[insert DOI]. Use of this Accepted Version is subject to the publisher’s Accepted Manuscript terms of use: https://www.springernature.com/gp/open-research/policies/accepted-manuscript-terms.

\end{abstract}
\section{Introduction \& Related work}
TSC is a key area in machine learning and signal processing, with applications such as stock market prediction \cite{Stock}, medical analysis \cite{Arrhythmia}, and climate forecasting \cite{weather}. Despite their success, deep neural networks (DNNs) for TSC remain vulnerable to adversarial examples \cite{goodfellow_explaining_adversarial_examples}. Adversarial attacks are classified as white-box or black-box: the former assumes full model access, enabling gradient-based methods, while the latter operates with limited information. White-box attacks, extensively studied in CV, are gaining attention in time series. Time series data is prone to signal distortions, malicious alterations, and environmental noise, making adversarial attacks particularly devastating.

\textbf{Definition 1.1}
Given a dataset \( D = \{(X_i, Y_i)\}_{i=1}^{n} \), \( X_i = (x_1, x_2, \ldots, x_k)^T \in \mathbb{R}^k \) represents a univariate time series of length \( k \), and \( Y_i \in [C] \) denotes the ground truth labels. Let $f$ be a TSC model. An adversarial example is defined as a perturbed input \( X' = X + \delta \), where \( \delta = (\delta_1, \delta_2, \ldots, \delta_k)^T \in \mathbb{R}^k \) is a small perturbation that maximizes the model loss \( L(f(X', \theta), Y) \). 

$\delta$ is computed using \eqref{perturbation}:

\begin{equation}
\delta = \text{argmax} \, L(f(X + \delta, \theta), Y), \quad \delta \in S,
\label{perturbation}
\end{equation}

where \( S \) denotes the constraint set, for example, \( S = \{\delta : \|\delta\|_2 \leq \epsilon\} \) specifies that the perturbation should lie within an \( \epsilon \)-radius \( k \)-dimensional ball. This operation is called clipping. Asadulla et al. adapted CV-based adversarial attacks such as FGSM, BIM, and PGD, to time series tasks without accounting for the unique characteristics of time series data \cite{galib2023susceptibility}. PGD, introduced by Madry et al. \cite{madry2017towards}, is an iterative white-box attack that refines perturbations over multiple gradient ascent steps while ensuring the adversarial example remains using clipping. Due to the characteristics of time series data, crafting stealthy attacks is significantly more challenging than for images \cite{ding2023blackbox}. Therefore, in practice, attack methods on time series data need to adapt to this nature. In our experiments, we set attack parameters to more closely mimic realistic scenarios.

Pialla et al. \cite{pialla2022smooth} and Chang et al. \cite{dong2023swap} innovated a "smooth attack" named 'GM' and a more effective and stealthy method called 'SWAP' respectively tailored for TSC models. Carlini et al. \cite{carlini2017towards} formulated the C\&W attack to minimize perturbation while ensuring misclassification, with \( c \) balancing the trade-off.

 AT aims to learn the adversarial perturbation pattern and train models to ignore the perturbation through suitable regularization \cite{pialla2022smooth}\cite{rathore2020untargeted}. It is considered one of the most powerful defenses against adversarial attacks \cite{akhtar2021advances} but often shows limited resistance to unseen adversarial attack strategies and requires intensive computational resources. Madry et al. \cite{madry2017towards} used adversarial examples generated by PGD attack to improve the performance of AT, referred to as PGD-based AT and this has become a baseline for most defense methods and formulated AT as a min-max optimization problem. The inner maximization seeks to identify the most adversarial perturbation for the model, whereas the outer minimization adjusts the model to become robust against this worst-case perturbation. 

Defensive Distillation (DD) is one of the most well-known adversarial defense methods except AT \cite{papernot2016distillation}. While some methods have successfully attacked DD \cite{carlini2017towards}, it remains a common baseline in the CV field \cite{strauss2018ensemble}. However, DD has yet to be applied in the time series domain. This technique strengthens the robustness of DNNs against adversarial attacks by raising the softmax layer’s temperature during the teacher model’s training, and then using the softened outputs to train the student model.

Zeng et al. introduced a lightweight defense method in CV based on data augmentation with randomness\cite{zeng2020data}. Iwana et al. highlighted various data augmentation methods for TSC with DNNs \cite{iwana2021empirical}. Pialla et al. further investigated the use of data augmentation techniques in TSC \cite{pialla2022data}. They demonstrated data augmentation can effectively enhance model accuracy and mitigate overfitting and also explored the benifits of ensembling these models. However, there is a research gap in lightweight defense methods for TSC. We address this gap by introducing our proposed methods inspired by data augmentation and ensemble. To facilitate reproducibility and future research, the source code of our implementation is publicly available at: \url{https://github.com/Yi126/Lightweight-Defence}. The main contributions of our research are summarized as follows:

\begin{itemize}
    \item[$\bullet$]
        Proposed five data augmentation-based defense methods and two combined methods for time series. One combined method improved both generalization and adversarial robustness of TSC models via ensemble learning, with training time reduced to 29.37\% of PGD-based AT on InceptionTime.  
    \item[$\bullet$]
        Theoretically and empirically validated the effectiveness of the proposed defense methods.
    \item[$\bullet$]
        This work demonstrated comprehensive benchmarking by comparing with AT and DD using the proposed ensemble methods with two TSC models faced six white-box gradient-based attacks on UCR datasets \cite{dau2019ucr}.
\end{itemize}

\section{Methodology}

\begin{figure}
    \centering
    \includegraphics[scale=0.35]{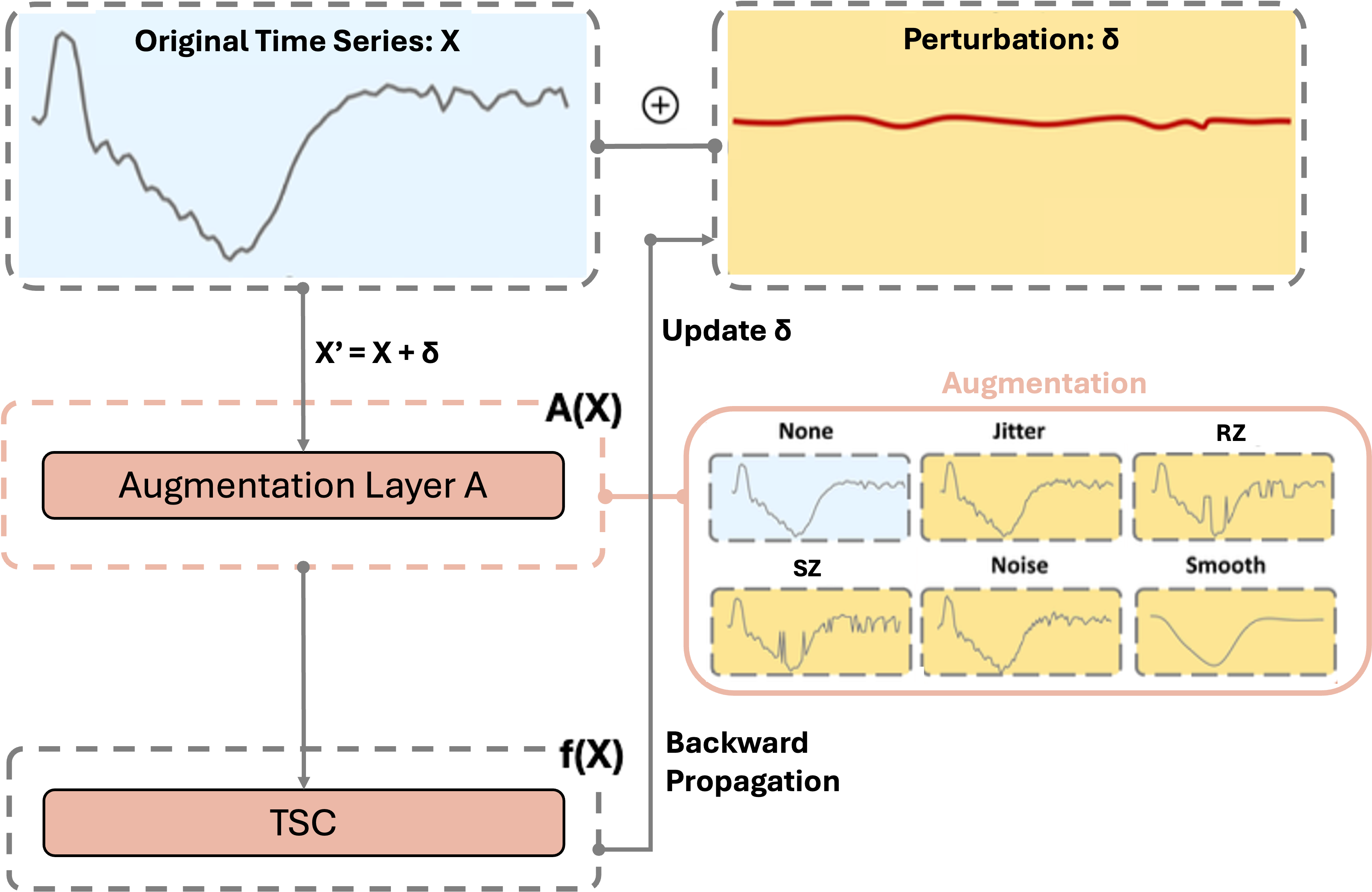}
    \caption{Schematic diagram of single data augmentation methods and SD.} \label{aug1}
\end{figure}

\subsection{Single Data Augmentation Methods (SDAMs)}

To improve the robustness of TSC models, five single data augmentation methods are proposed: Jitter, RandomZero, SegmentZero, Gaussian Noise, and Smooth Time Series. The implementation of each is shown in Fig.~\ref{aug1} in which corresponding method is selected every time the forward propagation happens. These methods enhance the diversity of time series data by applying data augmentation methods to the input time series before it was fed into the TSC model, thus making it harder for adversarial attacks to succeed. The detailed steps of these methods are provided in Algorithm \ref{alg:single}.

\textbf{Jitter}: This method generates a Bernoulli mask, creating noise based on Uniform distribution with a specified noise level, then adding the noise to the original data.

\textbf{RandomZero (RZ)}: This method involves masking random elements in the input data based on a Bernoulli distribution, then set the corresponding elements to zero.

\textbf{SegmentZero (SZ)}: This method sets segments of the input data to zero based on randomly chosen start timestamps and segment lengths which are the lengths of the masks. Then apply the masks to the data.

\textbf{Gaussian Noise (Noise)}: This method adds Gaussian noise to the input data. The noise is generated with a specified mean and standard deviation.

\textbf{Smooth Time Series (Smooth)}: This method smooths the time series data using a Gaussian kernel. The process involves constructing and normalizing a Gaussian kernel, and performing convolution with the time series data.

\begin{figure}
    \centering
    \includegraphics[scale=0.40]{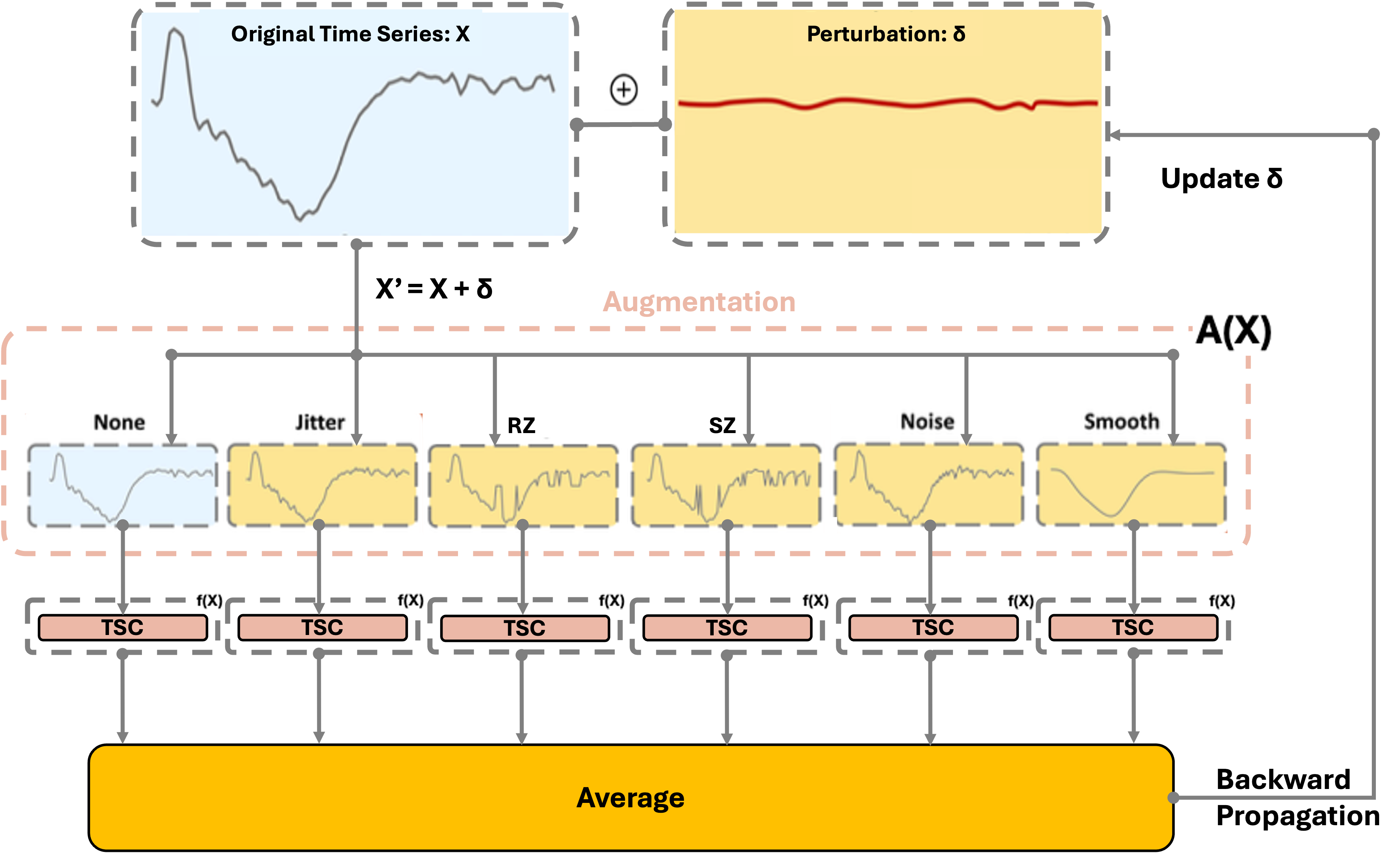}
    \caption{Schematic diagram of AD.} \label{ens}
\end{figure}

\subsection{Combined Data Augmentation Methods}

\subsubsection{Shuffle Defence (SD)}
SD also directly deploys the augmentation layer before the TSC, for every epoch of forward propagation, this augmentation layer will randomly select a data augmentation method from the five methods illustrated above or one other method -- not implementing any augmentation (None). SD increases the variation between each epoch, thus increasing randomness to disrupt the tracking of gradients in gradient-based attacks. As shown in Fig.~\ref{aug1}, the method of None is also included in the data augmentation methods to be selected. For every epoch, the original time series go through a randomly chosen method in the orange rectangle. 

\begin{algorithm}[H]
\caption{Single Data Augmentation Methods}
\label{alg:single}

\textbf{Input:} Input time series \( X \). 
\textbf{Output:} Augmented time series \( X' \). \\
\textbf{Hyperparameters:}

Jitter: noise level \(\text{noise\_level}\), probability \( p_j \). \\
RandomZero: probability \( p_r \). \\
SegmentZero: total length of the mask \(\text{total\_zero\_length}\), maximum length for each mask segment \(\text{max\_segment\_length}\). \\
Gaussian Noise: mean \( \mu \), standard deviation \(\sigma_g\). \\
Smooth Time Series: kernel size \(\text{kernel\_size}\), standard deviation \(\sigma_s\).

\textbf{Method 1: Jitter} \\
1. Generate a Bernoulli mask \( m_j \): $ \quad m_j \sim \text{Bernoulli}(p_j) $;

2. Generate uniformly sampled noise \( n_j \): $ \quad n_j \sim \text{Uniform}(-1, 1) \cdot \text{noise\_level} $;

3. Add noise to original data \( X' = X + m_j \odot n_j \).

\textbf{Method 2: RandomZero} \\
1. Generate a Bernoulli mask \( m_r \): $\quad m_r \sim \text{Bernoulli}(p_r)$;

2. Setting the corresponding elements to zero: \( X' = X \odot (1 - m_r) \).

\textbf{Method 3: SegmentZero} \\
1. Randomly determine segment lengths \( l_i \) and start positions \( s_i \) respectively;

2. Generate masks \( m_i \) for each segment:
\[
m_i[j] = \begin{cases} 
0, & \text{if } s_i \leq j \leq s_i + l_i; \\
1, & \text{otherwise}.
\end{cases}
\]
3. Combine masks: \( m = \bigcap_i m_i \); \\
4. Compute \( X' = X \odot m \).

\textbf{Method 4: Gaussian Noise} \\
1. Generate Gaussian noise \( n_g \): $\quad n_g \sim \mathcal{N}(\mu, \sigma_g^2)$;

2. Compute \( X' = X + n_g \).

\textbf{Method 5: Smooth Time Series} \\
1. Construct and normalize a one-dimensional Gaussian kernel \( ker \) with kernel size \(\text{kernel\_size}\), standard deviation \(\sigma_s\) and mean zero;

2. Convolve \( ker \) with \( X \): $ \quad X' = X \ast ker. $

Return \( X' \)
\end{algorithm}

\subsubsection{Average Defence (AD)}
Initially, the six methods of SD are applied separately to the same TSC model, and these six models are trained independently. During the testing phase, their outputs are averaged to derive the final classification result. This ensemble model approach mitigates errors made by individual base models.

As illustrated in Fig.~\ref{ens}, each base model within the ensemble generates predictions based on its own learned experiences. By employing averaging, AD enhances both the generalization capabilities and adversarial robustness of TSC models. For "easy examples," the predictions of the individual models largely converge. For "difficult examples," the predictions diverge, but on average, they tend to be closer to the correct answer.

\section{Theoretical analysis}

\textbf{Theorem 3.1}
By Definition 1, assume that $A_t$ is a data augmentation layer applied at the $t$-th epoch. Under the assumption that $\delta$ is small, the augmentation layer reduces the model’s sensitivity to perturbations from gradient-based attacks, resulting in improved robustness. 

\textbf{Proof of Theorem 3.1}
Given $\delta$ is sufficiently subtle compared to $X$, we assume that $A_t$ has a linear response to $\delta$, allowing us to ignore higher-order terms in the Taylor expansion of $A_t$ at $X$, $\nabla_X A_t$ is the Jacobian matrix of $A_t(X)$ evaluated at $X$:
\begin{equation}
f(A_t(X')) \approx f(A_t(X) + \nabla_X A_t \cdot \delta).
\label{taylor1}
\end{equation}
Given $\nabla_X A_t \cdot \delta$ is also diminutive compared to $A_t(X)$, we can perform a Taylor expansion of $f$ at $A_t(X)$, ignoring higher-order terms:
\begin{equation}
f(A_t(X')) \approx f(A_t(X)) + \nabla_X f(A_t(X)) \cdot \nabla_X A_t \cdot \delta.
\label{taylor2}
\end{equation}
$\delta$ at the $t$-th epoch can also be derived from gradient-based attacks which aims to maximize the model’s output deviation:
\begin{equation}
\delta = \frac{\nabla_X f(A_{t-1}(X))}{f(A_{t-1}(X))}.
\label{gradient}
\end{equation}
Since the attack relies on the gradient, $\delta$ is proportional to the input gradient $\nabla_X f$, and as shown in \eqref{perturbation}, clipping ensures that the perturbation remains subtle in scale. Thus, the scalar term in \eqref{gradient} becomes negligible. This results in the differentiation between the original and perturbed outputs as follows:
\begin{equation}
\begin{split}
f(A_t(X')) - f(A_t(X)) = &\ \nabla_X f(A_t(X)) \cdot \nabla_X f(A_{t-1}(X)) \\&\ \cdot \nabla_X A_t \cdot \delta.
\end{split}
\label{differentiation}
\end{equation}
The association between $\nabla_X f(A_{t-1}(X))$ and $\nabla_X f(A_t(X))$ introduces variability given the differences between $A_{t-1}(X)$ and $A_t(X)$. Furthermore, $\nabla_X A_t$ does not necessarily align with $\nabla_X f$ which substantially reduces their dot product. This results in a smaller output difference when data augmentation is applied. In contrast, without augmentation layer, the attack operates directly on $X$, and the output perturbation can be articulated as \eqref{noaug}:
\begin{equation}
f(X') - f(X) = \nabla_X f(X) \cdot \nabla_X f(X).
\label{noaug}
\end{equation}
Owing to the fact that they are the same vector, the output differences can be easily amplified, making the model more sensitive to small perturbations. Therefore, the introduction of the augmentation layer reduces the overall impact of input perturbations by introducing randomness and breaking the alignment between the input gradient and the perturbation. The interaction of gradients between augmentation layers at different epochs further reduces the propagation of perturbations, thereby improving the robustness of the model.

\textbf{Theorem 3.2}
When employing AD based on models with SDAMs, the classification accuracy of the ensemble model is better than that of the average of base models.

\textbf{Proof of Theorem 3.2}
Assume $N$ base models $h_1(X), \ldots, h_N(X)$, each with variance $\sigma_i^2$. Let $\bar{\sigma}^2 $ be the average variance and $\text{Cov}(h_i(X), h_j(X))$ be the covariance between base models. 
Since the ensemble's output is the average of the outputs of all base models, Using the properties of variance and covariance, we have the following expression:
\begin{equation}
\text{Var}(\bar{h}(X)) = \frac{1}{N^2} \left( \sum_{i=1}^{N} \sigma_i^2 + \sum_{i \neq j}^{N} \text{Cov}(h_i(X), h_j(X)) \right).
\label{varensemble}
\end{equation}
For independent base models, $\text{Cov}(h_i(x), h_j(x)) = 0$ for $i \neq j$. Then we have:
\begin{equation}
\text{Var}(\bar{h}(x)) = \frac{\bar{\sigma}^2}{N}.
\label{independent}
\end{equation}
In the completely correlated case, $\text{Cov}(h_i(x), h_j(x)) = \sigma_i \sigma_j$, thus:
\begin{equation}
\text{Var}(\bar{h}(x)) = \bar{\sigma}^2.
\label{correlated}
\end{equation}
AD is based on models with SDAMs which is trained with the same training set and TSC model, but with different random data augmentations. Thus, the base models are neither fully independent nor fully correlated. Therefore, the variance of the outputs when using AD will be lower than the average variance of these six base models.

Let the bias for each base model be $\mathbb{E}[h_i(X)] - y$, where $y$ is the true output. Then the bias of the ensemble model is: $ \frac{1}{N} \sum_{i=1}^{N} \mathbb{E}[h_i(X)] - y $ which equals the average bias of the base models.

The overall error of a model can be expressed using the bias-variance tradeoff \cite{tradeoff}:
\begin{equation}
\text{Error} = \text{Bias}^2 + \text{Variance} + \text{Irreducible Error}.
\label{tradeoff}
\end{equation}
The irreducible error is caused by inherent noise and unpredictability in the data and remains the same across all models for a given dataset. Thus, the irreducible error of the ensemble model matches the average irreducible error of the base models. Since the ensemble model having a lower variance than the average variance of the base models, its total error is reduced. This indicates that the classification accuracy of the TSC model with AD exceeds the average accuracy of models using SDAMs.

Moreover, Employing ensemble reduces the classification error rate facing adversarial attacks \cite{deng2024understanding}, as unaffected base models can compensate for those that are compromised, improving the model’s overall stability.

\section{Experiment setup}

\subsection{Dataset}
The UCR Archive 2018 were used during the experiments \cite{dau2019ucr}. It collects 128 time series related sub-datasets from various fields, featuring different lengths and numbers of categories, and has become a well-known benchmark in TSC.

\subsection{Model selection and performance metrics}
For different network structures, three models were selected, including InceptionTime \cite{inception} and ResNet18 \cite{resnet}. Natural accuracy (NA) and F1 score (F1) are chosen as the metrics to evaluate the classification performance. Robust accuracy (RA) is used to measure the robustness of the model against attacks. Training time (Time) is used to assess the computational resources and time required to deploy defense methods.

\subsection{Implementation Details }

The experiments were conducted on a computer equipped with an intel core i7 13700K, 64GB of memory and an NVIDIA RTX 2080ti GPU. 

For the attack phase, perturbations generated were constrained within the range of ±0.1. These perturbations were initialized randomly within a span of ±0.001. The attack methods we selected include all white-box attacks frequently used as baselines: FGSM, BIM, GM, SWAP, PGD, and C\&W. In FGSM, the step size for updates was 0.1; in BIM and PGD, the step size for each iteration was 0.0005; In SWAP, the difference between the largest and the second largest logits was 0.02. In C\&W optimization, the normalization parameter $c$ was set to $1 \times 10^{-5}$. These settings ensured the stealthiness of adversarial attacks in the field of time series. Both training and attack processes were run for 1000 epochs. 

For the defense phase, each of the employed techniques is detailed below:
\begin{itemize}
    \item Jitter: $p = 0.75$, noise\_level = 1
    \item RandomZero: $p = 0.5$
    \item SegmentZero: total\_zero\_length = 0.25, max\_segment\_length = 0.05
    \item Gaussian Noise: $\mu = 0$, $\sigma = 0.3$
    \item Smooth Time Series: kernel\_size = 10, $\sigma = 5$
\end{itemize}

In AT, the model was trained using both natural samples and adversarial samples generated through 40 iterations of the PGD method. In DD, to balance model performance and defense effectiveness, the temperature $T=10$ was chosen based on the experimental results from Papernot's paper \cite{papernot2016distillation}. 

Due to the randomness in the proposed defense method, we measured the NA and F1 five times each time and took the average value to obtain more accurate experimental results.

\section{Performance evaluation}

We trained and tested the two TSC models respectively, and compared their performance with AT, DD, None, SD and AD.

\begin{table}
\centering
\caption{Comparison between single data augmentation methods and None with Inceptiontime.}\label{tab1}
\setlength{\tabcolsep}{7pt} 
\renewcommand{\arraystretch}{1} 
\resizebox{0.9\textwidth}{!}{%
\begin{tabularx}{\textwidth}{|l|X|X|X|X|X|X|}
\hline
 & None & Jitter & RZ & SZ & Noise & Smooth \\
\hline
NA & 0.823 & 0.773 & 0.779 & 0.752 & 0.759 & 0.813 \\
F1 & 0.816 & 0.756 & 0.769 & 0.738 & 0.748 & 0.810 \\
Time & 78252 & 83729 & 82934 & 89265 & 83846 & 88760 \\
\hline
\end{tabularx}
}
\end{table}

Table~\ref{tab1} presents the results of each single data augmentation method and None with Inceptiontime. Compared to None, all other methods demonstrated a decrease in classification performance to various degrees. This outcome is expected, as all data augmentation methods altered the natural samples and introduced randomness. Even though the classifier and augmentation layer were trained together, they could not completely match the performance of the original classifier. Table~\ref{tab2} shows the performance of all defense methods with Inceptiontime, where a decrease in classification performance was observed for all methods except for AD. As discussed earlier, AD outperformed all single data augmentation methods and all baseline methods on UCR datasets in terms of natural accuracy and F1 score.

\begin{table}
\caption{Comparison between our defense methods and baselines with Inceptiontime.}\label{tab2}
\setlength{\tabcolsep}{5pt} 
\renewcommand{\arraystretch}{1} 
\resizebox{1\textwidth}{!}{%
\begin{tabularx}{\textwidth}{|l|X|X|X|X|X|}
\hline
 & None & AT & SD(Ours) & DD & AD(Ours) \\
\hline
NA & 0.823 & 0.807 & 0.621 & 0.808 & \textbf{0.839} \\
F1 & 0.816 & 0.800 & 0.593 & 0.799 & \textbf{0.832} \\
Time & 78252 & 1725338 & \textbf{98404} & 238894 & 506787 \\
RA(FGSM) & 0.486 & 0.579 & 0.589 & 0.480 & \textbf{0.629} \\
RA(BIM) & 0.453 & \textbf{0.577} & 0.539 & 0.483 & 0.543 \\
RA(GM) & 0.478 & 0.556 & 0.542 & \textbf{0.579} & 0.561 \\
RA(SWAP) & 0.463 & 0.509 & 0.573 & 0.580 & \textbf{0.581} \\
RA(PGD) & 0.437 & 0.538 & 0.543 & 0.457 & \textbf{0.559} \\
RA(C\&W) & 0.446 & 0.520 & 0.532 & 0.571 & \textbf{0.572} \\
\hline
\end{tabularx}
}
\end{table}

\begin{figure}[htbp]
    \centering
    \begin{subfigure}[b]{0.32\textwidth}
        \includegraphics[width=\textwidth]{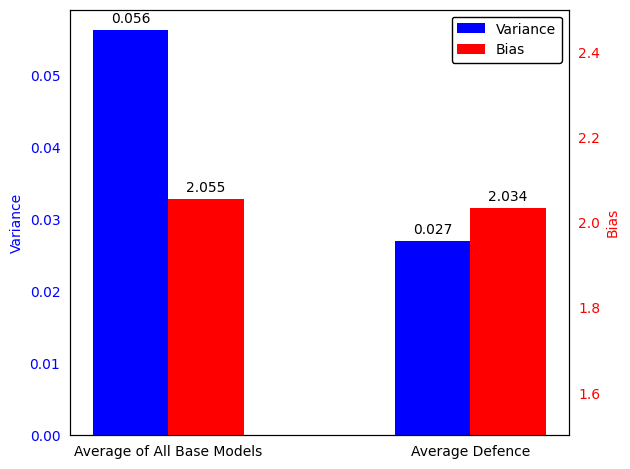}
        \caption{}
    \end{subfigure}
    \hfill
    \begin{subfigure}[b]{0.32\textwidth}
        \includegraphics[width=\textwidth]{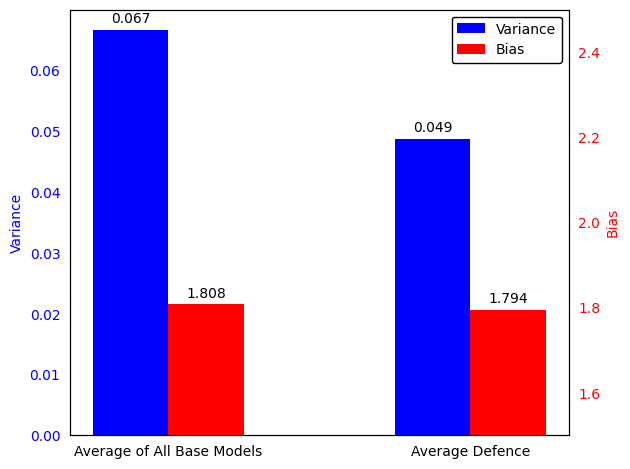}
        \caption{}
    \end{subfigure}
    \hfill
    \begin{subfigure}[b]{0.32\textwidth}
        \includegraphics[width=\textwidth]{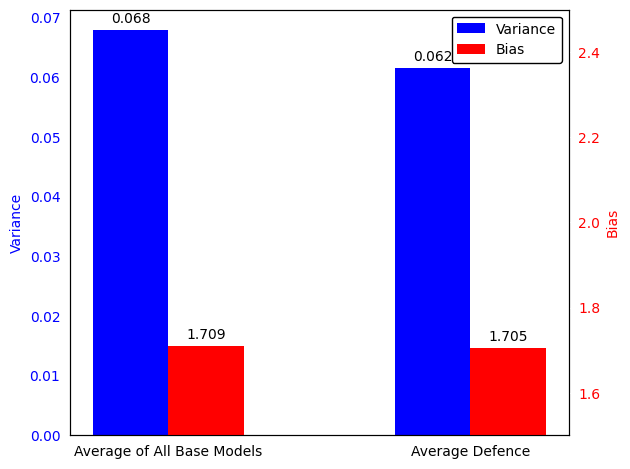}
        \caption{}
    \end{subfigure}

    \vspace{0.3cm}

    \begin{subfigure}[b]{0.32\textwidth}
        \includegraphics[width=\textwidth]{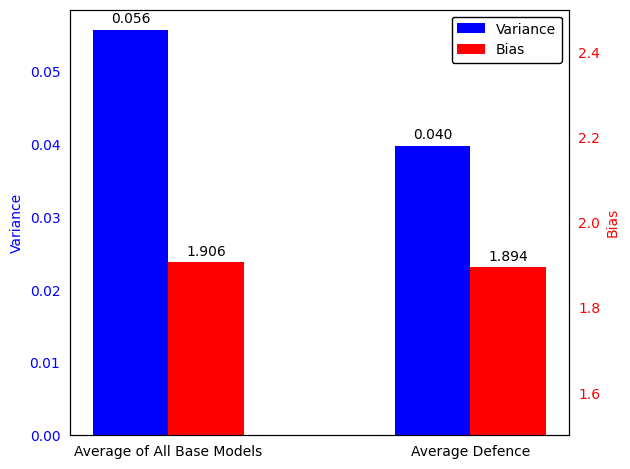}
        \caption{}
    \end{subfigure}
    \hfill
    \begin{subfigure}[b]{0.32\textwidth}
        \includegraphics[width=\textwidth]{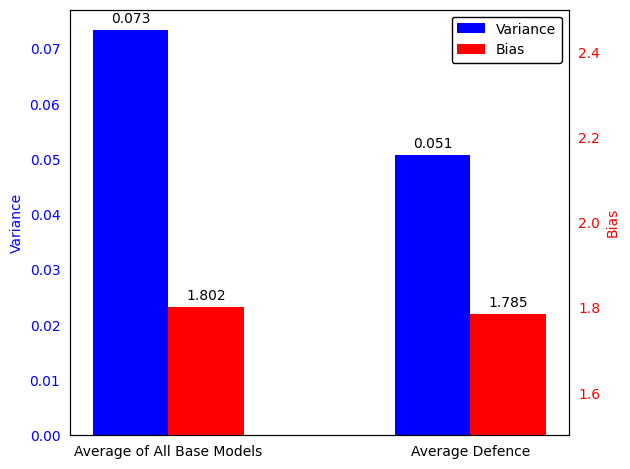}
        \caption{}
    \end{subfigure}
    \hfill
    \begin{subfigure}[b]{0.32\textwidth}
        \includegraphics[width=\textwidth]{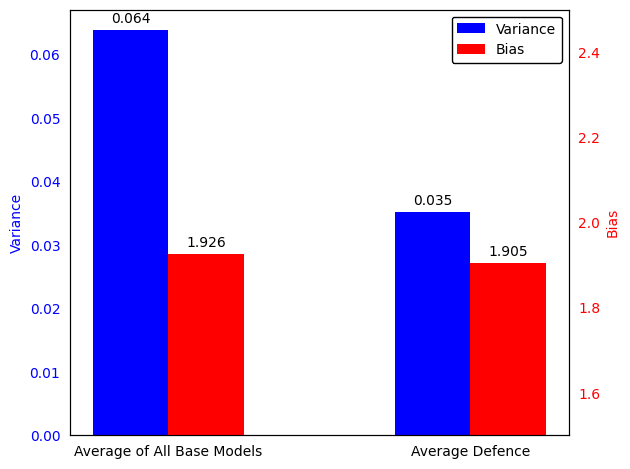}
        \caption{}
    \end{subfigure}

    \caption{Illustration of AD reduces variance. Each of the six graphs depicts the comparison between variance and bias of using AD and the average variance and bias of all base models under one sample respectively.}
    \label{var}
\end{figure}

We extracted six samples from the CricketX \cite{cricket} sub-dataset in UCR datasets, calculated the variance and bias of Inceptiontime’s outputs with AD, as well as the average variance and bias of all base models' outputs. As shown in Fig.~\ref{var}, AD lowers variance as expected by Theorem 3.2, thus improving the classification accuracy of natural samples.

Table~\ref{tab2} highlights AD’s superior robustness against adversarial attacks, with higher RA than other baselines in most cases. Additionally, AD required less than one-third of the computational time compared to AT, saving more resources than fast AT method \cite{jia2022boosting}. DD required only about half of computational time compared to AD and effectively defended against GM, C\&W, and SWAP attacks but was less effective against FGSM, BIM, and PGD. 

In the ResNet18 experiments, we used a subset of UCR datasets to save computational resources. To ensure data diversity and generalizability, we selected two sub-datasets from each UCR category.

According to Table~\ref{tab3}, AD consistently demonstrated strong defense performance on ResNet18, outperforming all baselines in adversarial robustness and improving NA by about 0.05. DD showed mediocre results, while AT performed steadily across both models but consumed significantly more computational resources without being as effective as AD. Although SD had the shortest Time among the defense methods, its generalization ability and robustness were very poor. 

\begin{table}
\caption{Comparison between our defense methods and baselines with ResNet18}\label{tab3}
\setlength{\tabcolsep}{5pt} 
\renewcommand{\arraystretch}{1} 
\resizebox{1\textwidth}{!}{%
\begin{tabularx}{\textwidth}{|l|X|X|X|X|X|}
\hline
 & None & AT & SD(Ours) & DD & AD(Ours) \\
\hline
NA & 0.807 & 0.800 & 0.618 & 0.795 & \textbf{0.856} \\
F1 & 0.803 & 0.796 & 0.595 & 0.790 & \textbf{0.852} \\
Time & 5751 & 220922 & \textbf{7097} & 12802 & 34916 \\
RA(FGSM) & 0.500 & 0.648 & 0.591 & 0.519 & \textbf{0.684} \\
RA(BIM) & 0.492 & 0.642 & 0.553 & 0.458 & \textbf{0.667} \\
RA(GM) & 0.491 & 0.641 & 0.552 & 0.467 & \textbf{0.665} \\
RA(SWAP) & 0.592 & 0.736 & 0.578 & 0.652 & \textbf{0.813} \\
RA(PGD) & 0.495 & 0.643 & 0.481 & 0.460 & \textbf{0.668} \\
RA(CW) & 0.489 & 0.638 & 0.477 & 0.456 & \textbf{0.663} \\
\hline
\end{tabularx}
}
\end{table}

\section{Conclusion, limitation and future work}

In this paper, we proposed lightweight methods to defend against adversarial attacks on TSC, demonstrating their effectiveness both theoretically and empirically. AD significantly improved generalization and adversarial robustness for mainstream models while using 29.37\% of the computational resources required by AT. Our work also serves as a benchmark, comparing AD with AT and DD on two TSC models under six white-box gradient-based attacks, making it a practical and efficient defense solution.

While our primary focus is on white-box robustness, future work could extend evaluations to black-box scenarios, further solidifying the method’s effectiveness. Additionally, although AD already achieves a strong balance between robustness and efficiency, exploring adaptive augmentation strategies or optimizing ensemble base models may further improve its performance. Investigating deployment in real-time or streaming TSC systems could also be a valuable direction, ensuring its applicability in latency-sensitive environments like edge computing and real-time anomaly detection. By demonstrating that effective adversarial defense can be achieved with significantly reduced computational cost, our work paves the way for more efficient and scalable defense mechanisms in TSC.

\bibliographystyle{splncs04}

\end{document}